\documentclass{article}

\usepackage{arxiv_ml_institute}

\usepackage[utf8]{inputenc} 
\usepackage[T1]{fontenc}    
\usepackage{hyperref}       
\usepackage{url}            
\usepackage{booktabs}       
\usepackage{amsfonts}       
\usepackage{nicefrac}       
\usepackage{microtype}      
\usepackage{xcolor}         

\usepackage{graphicx}
\usepackage{subfigure}

\usepackage{amsmath}
\usepackage{amssymb}
\usepackage{mathtools}
\usepackage{amsthm}
\usepackage{bm}
\usepackage{enumitem}

\usepackage{algorithm}
\usepackage{algpseudocode}

\usepackage{pifont}

\usepackage{tabularx}
\usepackage{multirow}
\usepackage{siunitx}
\usepackage{array}
\usepackage{makecell}
\newcolumntype{R}[1]{>{\raggedright\arraybackslash}p{#1}}
\newcolumntype{C}[1]{>{\centering\arraybackslash}p{#1}}
\newcolumntype{L}[1]{>{\raggedleft\arraybackslash}p{#1}}
\theoremstyle{plain}
\newtheorem{theorem}{Theorem}[section]

\theoremstyle{definition}
\newtheorem{definition}[theorem]{Definition}

\theoremstyle{remark}

\newcommand\Bg{\bm{g}}

\newcommand\Bp{\bm{p}}

\newcommand\Bx{\bm{x}}
\newcommand\By{\bm{y}}
\newcommand\Bz{\bm{z}}

\newcommand\BF{\bm{F}}

\newcommand\BI{\bm{I}}

\newcommand\BS{\bm{S}}

\newcommand\BW{\bm{W}}
\newcommand\BX{\bm{X}}
\newcommand\BY{\bm{Y}}
\newcommand\BZ{\bm{Z}}

\newcommand\Bxi{\bm{\xi}}

\newcommand\BXi{\bm{\Xi}}

 \newcommand{\dR}{\mathbb{R}}

\newcommand{\rE}{\mathrm{E}} 
 
 \newcommand{\rJ}{\mathrm{J}} 
 \newcommand{\rL}{\mathrm{L}}

\newcommand{\rS}{\mathrm{S}}

\newcommand\nn{\mathrm{new}}

\newcommand{\NRM}[1]{{{\left\| #1\right\|}}}

\newcommand{\soft}{\mathrm{softmax}}

\makeatletter
\newcommand{\dlmf}[1]{%
\citep[%
  \def\nextitem{\def\nextitem{, }}%
  \@for \el:=#1\do{\nextitem\href{http://dlmf.nist.gov/\el}{(\el)}}%
]{Olver:10}%
}
\makeatother

\definecolor{YColour}{RGB}{212, 101, 209}
\definecolor{RColour}{RGB}{140, 195, 98}
\definecolor{ZColour}{RGB}{46, 91, 207}
\definecolor{SColour}{RGB}{213, 228, 173}
\definecolor{IOColour}{RGB}{189, 46, 98}
\definecolor{RSColour}{RGB}{247, 206, 70}
\definecolor{REColour}{RGB}{243, 176, 138}
\definecolor{HColour}{RGB}{150, 204, 220}

\title{Hopular: Modern Hopfield Networks for \mbox{Tabular Data}}

\author{
    Bernhard Sch\"{a}fl\footnotemark[2]~$~^{,}$\thanks{Corresponding author: Bernhard Sch\"{a}fl <\href{mailto:schaefl@ml.jku.at}{schaefl@ml.jku.at}>} \quad
    Lukas Gruber\footnotemark[2] \quad
    Angela Bitto-Nemling\footnotemark[2]~$~^{,}$\footnotemark[3] \quad
    Sepp Hochreiter\footnotemark[2]~$~^{,}$\footnotemark[3]\\ \\
  \footnotemark[2]~~ELLIS Unit Linz and LIT AI Lab, Institute for Machine Learning,\\
                  ~~Johannes Kepler University Linz, Austria\\
  \footnotemark[3]~~Institute of Advanced Research in 
  Artificial Intelligence (IARAI)
}

\begin{document}

\maketitle

\begin{abstract}
    While Deep Learning excels in structured data as encountered in vision and
    natural language processing, it failed to meet its expectations on tabular data.
    For tabular data, Support Vector Machines (SVMs), Random Forests,
    and Gradient Boosting are the best performing techniques with Gradient Boosting in the lead.
    Recently, we saw a surge of Deep Learning methods that were tailored to tabular data
    but still underperform compared to Gradient Boosting on small-sized datasets.
    We suggest ``Hopular'', a novel Deep Learning architecture for medium- and
    small-sized datasets,
    where each layer is
    equipped with continuous modern Hopfield networks. The modern Hopfield networks
    use stored data to identify feature-feature, 
    feature-target, and sample-sample dependencies.
    Hopular's novelty is that every layer can directly access 
    the original input as well as the whole training set 
    via stored data in the Hopfield networks.
    Therefore, 
    Hopular can step-wise update its current model and the resulting 
    prediction at every layer 
    like standard iterative learning algorithms.
    In experiments on small-sized tabular datasets with less than 1,000 samples,
    Hopular surpasses Gradient Boosting, Random Forests, SVMs,
    and in particular several Deep Learning methods.
    In experiments on medium-sized tabular data with about 10,000 samples,
    Hopular outperforms XGBoost, CatBoost, LightGBM and 
    a state-of-the art Deep Learning method designed for tabular data.
    Thus, Hopular is a strong alternative to these methods on tabular data.
\end{abstract}

\section{Introduction}

Deep Learning has led to tremendous success
in vision and natural language processing, where it 
excelled on large image and text corpora \citep{LeCun:15,Schmidhuber:15}.
While it yielded competitive results on large tabular datasets \cite{Avati:18,Simm:18,Zhang:19,Mayr:18},
so far it could not convince on small tabular data.
However, in real-world settings, 
small tabular datasets with less than 10,000 samples are ubiquitous.
They are found in life sciences, when building a model for a certain disease
with a limited number of patients,
for bio-assays in drug design, or for the effect of environmental soil contamination.  
The same situation appears in most industrial applications, when
a company wants to predict customer behavior, to control processes, 
to optimize its logistics, to market new products, or to employ predictive maintenance.
The omnipresence of small tabular datasets can also be witnessed at Kaggle challenges.
On small-sized and medium-sized tabular datasets with less than 10,000 samples,
Support Vector Machines (SVMs) \citep{Boser:92,Cortes:95,Scholkopf:02},
Random Forests \citep{Ho:95,Breiman:01} and, in particular,
Gradient Boosting \citep{Friedman:01}
typically outperform Deep Learning methods with 
Gradient Boosting having the edge. 
In real world applications, 
the best performing and most prevalent Gradient Boosting variants are
XGBoost \citep{Chen:16},
CatBoost \citep{Dorogush:17,Prokhorenkova:18}, and
LightGBM \citep{Ke:17}.

Recently, research on extending Deep Learning
methods to tabular data has been intensified.
Some approaches to tabular data
are only remotely related to Deep Learning.
AutoGluon-Tabular stacks small neural networks for tabular
data \citep{Erickson:20}.
Neural Oblivious Decision Ensembles (NODE) generalizes
ensembles of oblivious decision trees by hierarchical representation
learning \citep{Popov:19}.
NODE is a hybrid of 
differentiable decision trees and neural networks.
DNF-Net builds neural structures corresponding to
logical Boolean formulas in disjunctive normal forms,
which enable localized decisions using small subsets of the features \citep{Abutbul:20}.

However, most research focused on adapting established Deep Learning techniques to
tabular data.
Modifications to deep neural networks 
like introducing leaky gates or skip connections
can improve their performance on tabular data \citep{Fiedler:21}. 
Even plain MLPs that are well-regularized work well on tabular data \citep{Kadra:21}.
Different regularization coefficients to each weight improve
the performance of Deep Learning architectures on tabular data \citep{Shavitt:18}.
TabularNet consists of three modules \citep{Du:21}.
First, it uses handcrafted cell-level feature extraction
with a language model for textual data.
Secondly, it uses both row and column-wise
pooling via bidirectional gated recurrent units.
Thirdly, a graph convolutional network captures
dependencies between cells of the table.

Many approaches that adapt Deep Learning methods to tabular data
use attention mechanisms from transformers \citep{Vaswani:17} and BERT \citep{Devlin:19}.
The TabTransformer learns contextual embeddings of categorical
features \citep{Huang:20}.
However, continuous features are not covered, therefore the feature-feature
interaction is limited.
The FT-Transformer maps
features to tokens that are fed into a transformer \citep{Gorishniy:21}.
The FT-Transformer performs well on tabular data but
all considered datasets have more than 10,000 samples.
TabNet uses an attentive transformer for sequential
attention to predict masked features \citep{Arik:21}.
Therefore, TabNet does instance-wise feature selection, that is,
can select the relevant features for each input differently.
TabNet also utilizes feature masking for pre-training, which was very successful
in natural language processing when pre-training the BERT model.
Also semi-supervised learning has been proposed for tabular
data using projections of the
features and contrastive learning \citep{Darabi:21}.
The contrastive loss is low if pairs of the same class have
high similarity.
Value Imputation and Mask Estimation (VIME) uses 
self- and semi-supervised learning of deep architectures for tabular
data \citep{Yoon:20}.
Like BERT, the network has 
to predict the values of the masked feature vectors, where the
target is always masked.
The success of BERT feature masking confirms that Deep Learning techniques
must employ strong regularization to be
successful on tabular data \citep{Kadra:21}.
A multi-head self-attentive neural network for modeling feature-feature interactions
was also used in AutoInt \citep{Song:19}.
So far we mentioned work, where attention mechanisms extract
feature-feature and feature-target relations. 
However, also inter-sample attention can be implemented, if the whole training
set is given at the input.
TabGNN uses a graph neural network for tabular data
to model inter-sample relations \citep{Guo:21}.
However, the authors focus on large tabular datasets with more than
40,000 samples.
SAINT contains both self-attention and inter-sample attention 
and embeds both categorical and continuous features
before feeding them into transformer modules \citep{Somepalli:21}.
SAINT uses self-supervised pre-training with a contrastive loss to
minimize the difference between original and mixed samples.
Non-Parametric Transformers (NPTs) also use feature self-attention 
and inter-sample attention \citep{Kossen:21}.
The feature self-attention identifies dependencies between
features, while inter-sample attention detects relations
between samples. 
As in previous approaches, BERT masking is used
during training, where
the masked feature values and the target have to be predicted.

We suggest \textbf{Hopular} to learn with modern \textbf{Hop}field networks from tab\textbf{ular} data.
Hopular is a Deep Learning architecture, where each layer is
equipped with continuous modern Hopfield networks
\citep{Ramsauer:21,Widrich:20nips}.
Continuous modern Hopfield networks can store two types of data:
(i) the whole training set or
(ii) the feature embedding vectors of the original input.
Like SAINT and NPT, Hopular can detect feature-feature, feature-target, 
sample-sample, and sample-target dependencies via modern Hopfield networks.
Hopular's novelty is that every layer can directly access 
the original input as well as the whole training set 
via stored data in the Hopfield networks. 
In each layer, the stored training set enables
similarity-, prototype-, or quantization-based learning methods like
nearest neighbor.
In each layer, the stored original input enables
the identification of dependencies between the features and the target.
Consequently, the current model and its prediction 
can be step-wise improved at every
layer via direct access to both the training set and the original input.
Therefore, a pass through a Hopular model is similar to standard
learning algorithms, which iteratively improve the current model and its prediction
by re-accessing the training set. The number of iterations
is fixed by the number of layers in the Hopular architecture.
As previous methods, Hopular uses a feature embedding and
BERT masking, where masked features have to be predicted.
Hopular is most closely related to SAINT \citep{Somepalli:21}
and Non-Parametric Transformers (NPTs) \citep{Kossen:21},
but in contrast to SAINT and NPTs, the whole training set and the original input
are provided via Hopfield networks at every layer and
not only at the input.

Recently, it was reported that Random Forests 
still outperform standard Deep Learning techniques on tabular datasets with up to 10,000
samples \citep{Xu:21}.
In \citep{ShwartzZiv:21}, the authors show that XGBoost
outperforms various Deep Learning methods that are designed for tabular data on
datasets that did not appear in the original papers.
Therefore, we test Hopular on exactly those datasets to see whether
it performs as well as XGBoost.
Furthermore, we test Hopular on
UCI datasets \citep{Ramsauer:21,Klambauer:17,Wainberg:16,Fernandez:14}. 
Hopular surpasses Gradient Boosting, Random Forests,
and SVMs but also state-of-the-art Deep Learning approaches
to tabular data~like~NPTs.

\section{Brief Review of Modern Hopfield Networks}\label{sec:MHN}

We briefly review  
continuous modern Hopfield networks.
Their main properties are that they retrieve 
stored patterns with only one update 
and that they have exponential storage capacity
\citep{Ramsauer:21}.

We assume a set of patterns $\{\Bx_1,\ldots,\Bx_N\} \subset \dR^d$
that are stacked as columns to 
the matrix $\BX = \left( \Bx_1,\ldots,\Bx_N \right)$ and a 
state pattern (query) $\Bxi \in \dR^d$ that represents the current state. 
The largest norm of a stored pattern is
$M = \max_{i} \NRM{\Bx_i}$.
Continuous modern Hopfield networks with state $\Bxi$
have the energy
\begin{equation}
\label{eq:energy}
\rE  \ =   \ -  \ \beta^{-1} \ \log \left( \sum_{i=1}^N
\exp(\beta \Bx_i^T \Bxi) \right) +  \ \beta^{-1} \log N  \ + \
\frac{1}{2} \ \Bxi^T \Bxi  \ +  \ \frac{1}{2} \ M^2 \ .
\end{equation}
For energy $\rE$ and state $\Bxi$, the update rule 
\begin{equation}
\label{eq:main_iterate}
\Bxi^{\nn} \ = \ f(\Bxi;\BX,\beta) = \ \BX \ \Bp = \   \BX \ \soft ( \beta \BX^T \Bxi)
\end{equation}
has been proven to converge globally  
to stationary points of the energy $\rE$, which are almost always local minima 
\citep{Ramsauer:21}.
The update rule Eq.~\eqref{eq:main_iterate}
is also the formula of the well-known transformer attention mechanism
\citep{Vaswani:17,Ramsauer:21}, therefore Hopfield retrieval and
transformer attention coincide.

The {\em separation} $\Delta_i$  of a 
pattern $\Bx_i$ is defined as its minimal dot product difference to any of the other 
patterns:
$\Delta_i = \min_{j,j \not= i} \left( \Bx_i^T \Bx_i - \Bx_i^T \Bx_j \right)$. 
A pattern is {\em well-separated} from the data if $
 \Delta_i  \geq \nicefrac{2}{\beta N} + \nicefrac{1}{\beta} \log \left( 2 (N-1)  N  \beta  M^2 \right)$.
If the patterns $\Bx_i$ are well separated, the iterate Eq.~\eqref{eq:main_iterate}
converges to a fixed point close to a stored pattern.
If some patterns are similar to one another and, therefore, not well separated, 
the update rule Eq.~\eqref{eq:main_iterate} converges to 
a fixed point close to the mean of the similar patterns. 
This fixed point is a {\em metastable state} of the energy function
and averages over similar patterns.

The next theorem states that the update rule Eq.~\eqref{eq:main_iterate} typically converges after
one update if the patterns are well separated. Furthermore, it states
that the retrieval error is 
exponentially small in the separation $\Delta_i$ (for the proof see~\mbox{\citep{Ramsauer:21}}):
\begin{theorem}
\label{th:oneUpdate}
With query $\Bxi$, after one update the distance of the new point $f(\Bxi)$
to the fixed point $\Bx_i^*$ is exponentially small in the separation $\Delta_i$.
The precise bounds using the Jacobian $\rJ = \nicefrac{\partial
  f(\Bxi)}{\partial \Bxi}$ and its value $\rJ^m$ in the mean value
theorem are:
\begin{equation}
  \NRM{f(\Bxi) \ - \ \Bx_i^*}
  \ \leq \  \NRM{\rJ^m}_2 \ \NRM{\Bxi \ - \ \Bx_i^*}  \ ,
\end{equation}
\begin{equation}
  \NRM{\rJ^m}_2  \ \leq \
  2 \ \beta \ N \ M^2 \ (N-1) \ \exp(- \ \beta \
  (\Delta_i \ - \ 2 \
  \max \{ \NRM{\Bxi  \ - \ \Bx_i} , \NRM{\Bx_i^* \ - \ \Bx_i} \}  \ M) )\ .
\end{equation}
For given $\epsilon$ and 
sufficiently large $\Delta_i$, we have $\NRM{f(\Bxi) \ - \ \Bx_i^*} < \epsilon$,
that is, retrieval with one update.
The retrieval error $\NRM{f(\Bxi) \ - \ \Bx_i}$ of pattern $\Bx_i$
is bounded by
\begin{equation}
   \NRM{f(\Bxi) \ - \ \Bx_i} \ \leq \ 2 \ (N-1) \ \exp(- \ \beta \ (\Delta_i \ - \ 2 \ \max \{ \NRM{\Bxi  \ - \ \Bx_i} , \NRM{\Bx_i^* \ - \ \Bx_i} \} \ M) )  \ M  \ .
\end{equation}
\end{theorem}

The main requirement to modern Hopfield networks to
be suited for tabular data is that they can store and retrieve enough patterns.
We want to store a potentially large training set in every layer
of a Deep Learning architecture.
We first define what we mean by storing and retrieving patterns
from a modern Hopfield network.
\begin{definition}[Pattern Stored and Retrieved]
We assume that around every pattern $\Bx_i$ a sphere $\rS_i$ is given.
We say $\Bx_i$ {\em is stored} if there is a single fixed point $\Bx_i^* \in \rS_i$ to
which all points $\Bxi \in \rS_i$ converge,
and  $\rS_i \cap \rS_j = \emptyset$ for $i \not= j$.
We say $\Bx_i$ {\em is retrieved} for a given $\epsilon$ if 
iteration (update rule) Eq.~\eqref{eq:main_iterate} gives
a point $\tilde{\Bx}_i$ that is at least 
$\epsilon$-close to the single fixed point $\Bx_i^* \in \rS_i$. 
The retrieval error is $\NRM{\tilde{\Bx}_i - \Bx_i}$.
\end{definition}

\begin{figure}[b]
    \centering
    \includegraphics[width=\textwidth]{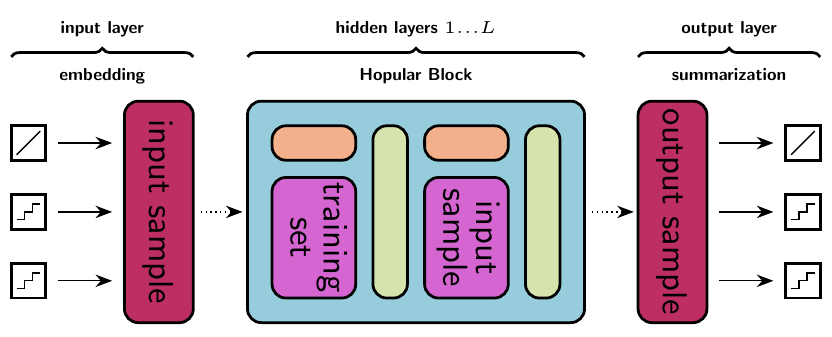}
    \caption{Architecture overview of Hopular. Hopular consists of three different types of layers or blocks. \textbf{(I) Embedding Layer}---each attribute of an original input sample is represented in an $e$-dimensional space. The original input sample itself is then represented by the concatenation of all of its attribute representations. \textbf{(II) Hopular Block}---the input representation is then refined by $L$ consecutive Hopular blocks. This is achieved by applying the two Hopfield modules $H_{s}$ and $H_{f}$ in an alternating way. \textbf{(III) Summarization Layer}---lastly, this refined current prediction is summarized by an attribute-wise mapping, leading to the final prediction.}
    \label{fig:hopular_overview}
\end{figure}

As with classical Hopfield networks, we consider patterns on the sphere, 
i.e.\ patterns with a fixed norm. 
For randomly chosen patterns, the number of patterns that can be stored
is exponential in the dimension $d$ of the space of the patterns 
(for the proof see \citep{Ramsauer:21}):
\begin{theorem}
\label{th:storage}
We assume a failure probability $0<p\leq 1$ and randomly chosen patterns 
on the sphere with radius $M:=K \sqrt{d-1}$. 
We define $a := \nicefrac{2}{d-1}  (1 + \ln(2 \beta K^2 p (d-1)))$, 
$b := \nicefrac{2  K^2  \beta}{5}$,
and $c:= \nicefrac{b}{W_0(\exp(a + \ln(b))}$,
where $W_0$ is the upper branch of the Lambert $W$ function \dlmf{4.13},
and ensure $c \geq \left( \nicefrac{2}{ \sqrt{p}}\right)^{\nicefrac{4}{d-1}}$.
Then with probability $1-p$, the number of random patterns 
that can be stored is: 
\begin{align}\label{eq:CapacityM}
    N \ &\geq \ \sqrt{p} \ c^{\frac{d-1}{4}}  \ .
\end{align}
Therefore it is proven for $c\geq 3.1546$ with
$\beta=1$, $K=3$, $d= 20$ and $p=0.001$ ($a + \ln(b)>1.27$)
and proven for $c\geq 1.3718$ with $\beta = 1$, $K=1$, $d = 75$, and $p=0.001$
($a + \ln(b)<{-0.94}$).
\end{theorem}

This theorem motivates to use continuous modern Hopfield networks
for tabular data, where we want to store the training set in each
layer of a Deep Learning architecture.
Even for hundreds of thousands of training samples, the 
continuous modern Hopfield network is able to store the training set
if the dimension of the pattern is large enough.

\begin{figure}[ht]
    \centering
    \includegraphics[width=\textwidth]{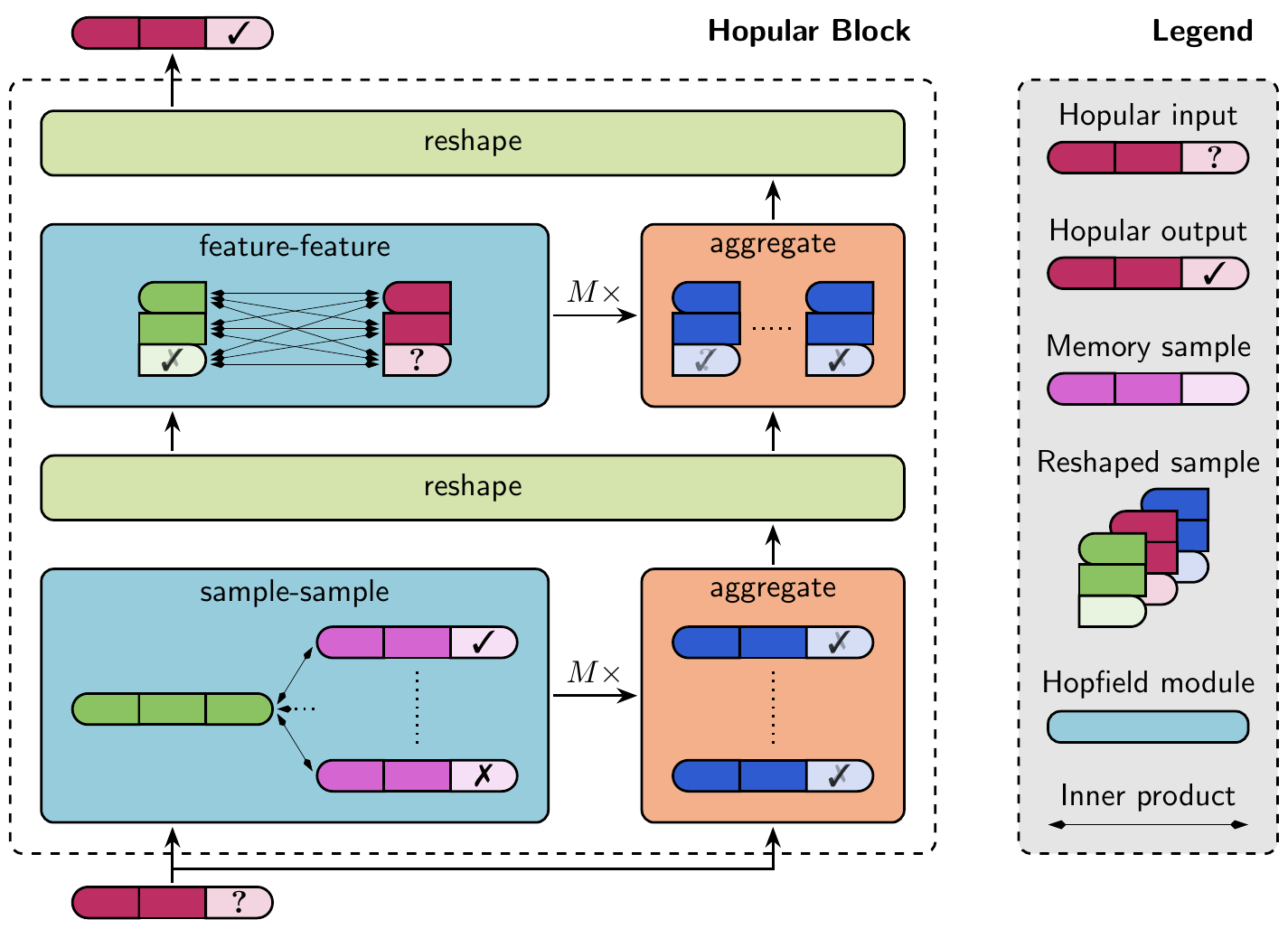}
    \caption{A Hopular Block. The first Hopfield module stores the whole training set and identifies sample-sample relations. The second Hopfield module stores the embedded input features and extracts feature-feature and feature-target relations. The Hopfield
    modules refine the current prediction by combining the aggregated retrievals
    of the $M$ Hopfield networks with their respective input.
    \label{fig:architecture_overview}}
\end{figure}

\section{Hopular: Modern Hopfield Networks for Tabular Data} \label{sec:mhn_tabular}

{\bf Hopular architecture.}
The Hopular architecture consists of an Embedding layer, several stacked 
Hopular blocks, and a Summarization layer as depicted in Figure~\ref{fig:hopular_overview}. As Hopular operates on features as well as on targets, we more generally refer to them as \emph{attributes}.

(i) The input to the Embedding Layer is an original input sample with $d$ attributes, 
including a masked target.
Categorical attributes are encoded as one-hot vectors, 
whereas continuous attributes are normalized to zero mean 
and unit variance.
Then a mapping to an \mbox{$e$-dimensional} embedding space is applied. 
The index of an attribute w.r.t.\ the position inside 
the sample as well as the attribute type are conserved 
by separate $e$-dimensional learnable embeddings. 
All three embedding vectors are element-wise summed and
serve as the final representation of an input attribute. 
The original input sample is then represented by the concatenation of all
attribute representations. This concatenation also initializes the current 
prediction vector $\Bxi \in \dR^{d \cdot e}$~\---{}~see
Figure~\ref{fig:hopular_embedding} of the Appendix.

(ii) The current prediction vector serves as input to a Hopular Block.
A Hopular block consecutively applies two different Hopfield modules.
Each of these Hopfield modules refines the current prediction vector by updating
the current predictions for all attributes and combining it with its
input via a residual connection.
Thus, in addition to the target, also the features of the original input sample must be predicted during training.
\figurename~\ref{fig:architecture_overview} illustrates the forward-pass 
of a single original input sample with the masked target  
indicated by the question mark (\textbf{?}).
All current attribute predictions are refined.
The masked target is transformed by the Hopular block 
to a corresponding prediction as indicated by a check mark (\ding{51}).
Also feature representations can be masked as with
BERT pre-training.

(iii) The Summarization Layer summarizes the refined current prediction vector resulting from the stacked Hopular blocks.
The current prediction vector is mapped to the final prediction vector by separately mapping each current feature prediction to the corresponding final prediction as well as mapping the current target prediction to the final target prediction~\---{}~see Figure~\ref{fig:hopular_summarization} of the Appendix. In the following we describe the components (I)--(II) of a Hopular Block.

{\bf (I) Hopfield Module $H_{s}$.} 
The first Hopfield module $H_{s}$ implements a modern Hopfield network for Deep Learning architectures
similar to {\tt HopfieldLayer} \citep{Ramsauer:21,Ramsauer:20} 
with the training set as fixed stored patterns.
The current input $\Bxi$ (which is also the current prediction from the previous
layer) to Hopfield module $H_{s}$
is interacting with the whole training data 
as described in Eq.~\eqref{eq:Hs}.
This is the update rule of 
continuous modern Hopfield networks as given in Eq.~\eqref{eq:main_iterate}.
Hence, the Hopfield module $H_{s}$ identifies sample-sample relations
and can perform similarity searches like a nearest-neighbor search 
in the whole training data.
$H_{s}$ can also average over training data that 
are similar to a mapping of the current prediction vector $\Bxi$.

Next, we describe Hopfield Module $H_{s}$ in more detail.
Let $d$ be the number of attributes, 
$e$ the embedding dimension of each single attribute, 
$h$ the dimension of the Hopfield embedding space, and $n$ the number of samples in the training set. 
The forward-pass for module $H_{s}$ with one Hopfield network and
current prediction vector $\Bxi \in \dR^{d \cdot e}$, 
learned weight matrices $\BW_{\Bxi},\BW_{\BX} \in \dR^{h \times (d \cdot e)}$, \ $\BW_{\BS} \in \dR^{(d \cdot e) \times h}$, the stored training set $\BX \in \dR^{(d \cdot e) \times n}$, and a fixed scaling parameter $\beta$
is given~as
\begin{equation}
\label{eq:Hs}
    H_s\left( \Bxi \right) \ = \ \BW_{\BS} \  \BW_{\BX} \ \BX \soft (\beta \ \BX^{T} \ \BW_{\BX}^{T} \ \BW_{\Bxi} \ \Bxi )\ .
\end{equation}
The hyperparameter $\beta$ allows to steer the type of fixed point 
the update rule Eq.~\eqref{eq:main_iterate} converges to, 
hence it may further amplify the nearest-neighbor-lookup of the sample-sample Hopfield module $H_{s}$. $H_{s}$ may contain more than one 
continuous modern Hopfield network.
In this case, the respective results are combined and projected, 
serving as the modules final output. 
We have $M$ separate Hopfield networks $H_{s}^{i}$, where the module output is defined as
\begin{align}\label{eq:Hs_combined}
    H_{s} \left(\Bxi \right) \ &= \ \BW_{G} \ \left( H_{s}^{1} \left( \Bxi \right)^{T}, \ldots,\ H_{s}^{M} \left( \Bxi \right)^{T} \right)^{T}\ ,
\end{align}
with vector $\left( H_{s}^{1} \left( \Bxi \right)^{T}, \ldots,\ H_{s}^{M} \left( \Bxi \right)^{T} \right)^{T}$
and a learnable weight matrix $\BW_{G} \in \dR^{(d\cdot{}e)\times{}(M\cdot{}d\cdot{}e)}$.

{\bf (II) Hopfield Module $H_{f}$.} 
The second Hopfield module $H_{f}$ implements a modern Hopfield network for Deep Learning architectures
via the layer {\tt Hopfield} \citep{Ramsauer:21,Ramsauer:20} 
with the embedded features of the original input sample as stored patterns.
The refined prediction vector from the previous layer
is reshaped and transposed
to the matrix $\BXi$, which serves as input to
the Hopfield module $H_{f}$.
$\BXi$ interacts with the embedded features
of the original input sample
as described in Eq.~\eqref{eq:Hf}.
Again, this is the update rule of 
continuous modern Hopfield networks as given in Eq.~\eqref{eq:main_iterate}.
Therefore, the Hopfield module $H_{f}$ extracts and models
feature-feature and feature-target relations.
Current feature and target predictions are adjusted and refined after they are associated with the original input sample feature representations.

Next, we describe Hopfield Module $H_{f}$ in more detail.
The matrix $\BXi \in \dR^{e \times d}$ is a transposed and reshaped version 
of current prediction vector $\Bxi$ with respect to the embedding dimension $e$.
Using the learned weight matrices $\BW_{\BXi},\BW_{\BY} \in \dR^{h \times e}$,  \
$\BW_{\BF} \in \dR^{e \times h}$, 
the embedded original input sample $\BY \in \dR^{e \times d}$, and a fixed scaling parameter $\beta$
the forward-pass is
\begin{equation}
\label{eq:Hf}
    H_f \left(\BXi \right) \ = \ \BW_{\BF} \ \BW_{\BY} \ \BY \soft \left( \beta \ \BY^{T} \ \BW_{\BY}^{T} \ \BW_{\BXi} \ \BXi \right).
\end{equation}
$H_{f}$ may contain more than one continuous modern Hopfield network,
which leads to an analog equation as~Eq.~\eqref{eq:Hs_combined} for~$H_{s}$.

{\bf Hopular architecture and Modern Hopfield Networks.}
Deep Learning could not convince so far on small tabular datasets, 
on the other hand iterative learning algorithms, 
like Gradient Boosting methods, are the best-performing methods in this domain. 
Therefore, we introduce a DL architecture that is able to mimic and 
extend these iterative algorithms by reaccessing the whole training set and 
refining the current prediction in each layer. 
Modern Hopfield Networks directly access an external memory
in a content-based fashion as depicted in Eq.~\eqref{eq:main_iterate}.
Hopular populates this external memory in two different ways: (a) Hopular uses the training set as an external memory, and (b) Hopular uses the embedded feature representations of the original input sample as external memory.
During training, retrieval from the respective memory is learned whereas
the type of fixed point of the modern Hopfield network, 
as described in Section~\ref{sec:MHN}, specifies the type of retrieved pattern.
Additionally, modern Hopfield networks
can retrieve patterns with only one update~\---{}~see Theorem~\ref{th:oneUpdate}.

Furthermore, their exponential storage capacity (Theorem~\ref{th:storage}) makes it possible to
retrieve patterns from external memories with even hundreds of thousands
instances.
Because of these properties Hopular can mimic iterative learning algorithms e.g.\ such based on gradient descent, boosting, or feature selection that refine the current prediction by re-accessing the training set in contrast to other Deep Learning methods for tabular data. Both NPTs and SAINT consider feature-feature and sample-sample interactions via their respective attention mechanisms which solely use the result of the previous layer. In contrast, Hopular not only uses the result of the previous layer but also the original input sample and the whole training set.
For example, our method can implement gradient boosting with a boosting step at each layer.
The ability to mimic iterative learning algorithms that are known to perform specifically well on tabular data makes modern Hopfield networks a promising
approach for processing tabular data.
For the instantiation variant that we use for our experiments
the Hopfield module $H_{s}$ identifies sample-sample relations
and can perform similarity searches like a nearest-neighbor search 
in the whole training data.
In the Appendix in Section~\ref{sec:iter_learn}
we give further intuition of how Hopular
can mimic iterative learning algorithms on the basis of two examples.

{\bf Hopular's Objective and Training Method.}
Hopular's objective is a weighted sum of 
the self-supervised loss for predicting masked features and
the standard supervised target loss.
In the following we explain the feature masking as well as the objective in more detail.

{\em Feature Masking.} 
We follow state-of-the-art Deep Learning methods 
like SAINT \citep{Somepalli:21}
and Non-Parametric Transformers (NPTs) \citep{Kossen:21}
that are tailored to tabular data and 
use BERT masking \citep{Devlin:19} of the input features.
Masked input features must be predicted during training.
Feature masking is an especially beneficial self-supervised approach when 
handling small datasets as it exerts a 
strong regularizing effect on the training procedure. 
The amount of masked features during training is 
determined by the masking probability, which is a hyperparameter of the model.
In Hopular, both features and targets can be masked during training,
while for inference only the target is masked.

{\em Objective.}
Hopular's objective is a weighted sum of the masked 
feature loss $\rL_f$ 
and the supervised target loss $\rL_t$.
The overall loss $\rL$ is
\begin{align}\label{eq:total_loss}
    \rL \ &= \ \gamma \ \rL_f \ + \  (1 \ - \ \gamma) \rL_t \ ,
\end{align}
where $\rL_t$ and $\rL_f$ are the negative logloss 
in case of discrete attributes and the mean squared error 
in case of continuous attributes with $\gamma$ as a hyperparameter. 
In our default hyperparameter setting $\gamma$ is
annealed using a cosine scheduler starting at $1$ with a final value of $0$.
Another essential hyperparameter for Hopular is $\beta$ 
in Eq.~\eqref{eq:Hs} and Eq.~\eqref{eq:Hf}.
A small $\beta$ retrieves a pattern close to the mean of
the stored patterns, while a large $\beta$ 
retrieves the stored pattern 
that is closest to the initial state pattern \citep{Ramsauer:21}.
For module $H_{s}$ a large $\beta$ value 
emphasizes a nearest-neighbor lookup mechanics.
For module $H_{f}$ a large $\beta$ value leads to less 
diluted features.
Thus, large $\beta$ values seem to be beneficial for Hopular.
Experiments confirm this assumption 
(see Section~\ref{sec:experiments}).

{\bf Hopular Pseudocode}. Algorithm~\ref{alg:Hopular} shows the forward pass of Hopular for
an original input sample~$\Bx$.

\begin{algorithm}[h]
   \caption{Forward pass of Hopular}
   \label{alg:Hopular}
\begin{algorithmic}[1]
   \Require Hopfield modules $H_s$ and $H_f$, embedding layer $E$,
   summarization layer $S$, number of features $d$, 
    number of Hopular blocks $L$
   and original input sample $\Bx \in \dR^d$
   \State $\Bx \gets \text{Mask}(\Bx)$
   \State $\Bxi \gets E(\Bx)$
   \For{$i=1$ {\bfseries to} $L$}
   \State $\Bxi \gets \Bxi + H_s(\Bxi)$
   \State $\BXi \gets \text{Reshape}(\Bxi^{T})$
   \State $\BXi \gets \BXi + H_f(\BXi)$
   \State $\Bxi \gets \text{Reshape}(\BXi)^{T}$
   \EndFor
   \State $\Bxi \gets S(\Bxi)$
\end{algorithmic}
\end{algorithm}

\section{Experiments}
\label{sec:experiments}

Since Deep Learning methods have already been successfully applied to larger tabular datasets \citep{Avati:18,Simm:18,Zhang:19,Mayr:18}
we want to know whether Hopular 
is competitive on small tabular datasets. 
In particular, we compare Hopular to XGBoost, CatBoost, LightGBM, and NPTs \citep{Kossen:21}.
Gradient Boosting has the lead on tabular data when excluding Deep Learning methods.
NPTs represent state-of-the-art Deep Learning methods
for tabular data,
as NPTs yielded very good results on small tabular datasets.

\subsection{Small-Sized Tabular Datasets}
\label{sec:experiments_uci}

In these experiments, we compare Hopular to other Deep Learning methods, XGBoost, CatBoost, and LightGBM on small-sized tabular datasets.

{\bf Methods Compared.}
We compare Hopular, XGBoost, CatBoost, LightGBM, NPTs, and other 24 machine learning
methods as described in \citep{Klambauer:17}.
The compared methods include 10 Deep Learning (DL) approaches.
Following \citep{Klambauer:17,Wainberg:16},
17 methods are selected from their respective method group 
as the model with the median performance over all datasets within each method group.
NPTs are used in a non-transductive setting for a fair comparison. 

{\bf Hyperparameter Selection.} 
All hyperparameters are selected on seperate validation sets. For NPTs we perform hyperparameter search as in Table~\ref{tab:hyp_npt}. This includes the hyperparameters that have already been successfully used in \citep{Kossen:21} on small- and medium-sized tabular datasets. 
This selection also serves as a constraint on the computational resources invested for Hopular.
For XGBoost, CatBoost, and LightGBM, we apply the same Bayesian hyperparameter optimization 
procedure as described in~\citep{ShwartzZiv:21}. For LightGBM we use the default hyperparameter ranges as specified by \texttt{hyperopt-sklearn}~\citep{Komer:14}.
Section~\ref{sec:hyperparameter_selection} of the Appendix describes the hyperparameter selection in more detail.

{\bf Datasets.} 
Following \citep{Klambauer:17}, 
we consider UCI machine learning repository datasets 
with less than or equal to 1,000 samples as being {\em small}.
We select 21 of these datasets and give an overview in Table~\ref{tab:small-sized datasets}.
The datasets themselves as well as the train/test splits are taken from~\citep{Fernandez:14}.
A detailed explanation of the dataset selection process as well as a description of the datasets can be found in Section~\ref{sec:dataset_explanation} of the Appendix.

\begin{table}[ht]
    \caption{Median rank of compared methods across 
    the datasets of the UCI machine learning repository. 
    Methods are ranked for each dataset according to the accuracy on the respective test set.
    Hopular achieves the lowest median rank of $7.5$, therefore is the best
    performing method across the considered UCI datasets. The complete list can be seen in Table~\ref{tab:uci_experiments_complete} of the Appendix.\label{tab:uci_experiments}}
    \begin{center}
        \begin{tabular}{lS[table-format=2.1]lS[table-format=2.1]}
            {\bf Method} & {\bf Rank} & {\bf Method} & {\bf Rank} \\
            \toprule
            Hopular (DL)                     &  7.5                             & CatBoost                            & 14.0 \\
            {\multirow{2}{*}{\quad $\vdots{}$}} & {\multirow{2}{*}{$\vdots{}$}} & LightGBM                            & 14.5 \\
                                          &                                     & {\multirow{2}{*}{\quad $\vdots{}$}} & {\multirow{2}{*}{$\vdots{}$}}\\
            Non-Parametric Transformers (DL) & 11.0                             &                                     &  \\
            XGBoost                          & 12.0                             & Stacking (Wolpert)                  & 28.0
        \end{tabular}
    \end{center}
\end{table}

{\bf Results.} Table~\ref{tab:uci_experiments} shows the median rank of all compared methods across 
the datasets of the UCI machine learning repository
(see Table~\ref{tab:uci_experiments_complete} of the Appendix for the complete list).
Methods are ranked for each dataset according to the accuracy on the respective test set.
17 method groups have been compared previously \citep{Wainberg:16}, to which
we add XGBoost \citep{Chen:16}, CatBoost \citep{Dorogush:17,Prokhorenkova:18}, LightGBM \citep{Ke:17},
NPTs \citep{Kossen:21}, Self-Normalizing Networks \citep{Klambauer:17}, and our Hopular.
Deep Learning methods are indicated by ``(DL)'' and are not grouped.
Hopular has a median rank of $7.5$, followed by Support Vector Machines with $9.5$, 
while NPTs, XGBoost, CatBoost, and LightGBM
have a median rank of $11$, $12$, $14$, and $14.5$ respectively.
Hopular with modern Hopfield networks as memory performs better than
other Deep Learning methods 
and in particular better than the closely-related NPTs.
{\bf Across the considered UCI datasets,
Hopular is the best performing method.} 

\subsection{Medium-Sized Tabular Datasets}
\label{sec:experiments_tabular}

In these experiments, we compare Hopular to other Deep Learning methods,
XGBoost, CatBoost, and LightGBM on medium-sized tabular datasets.
In \citep{ShwartzZiv:21}, the authors show that XGBoost outperforms various
Deep Learning methods that are designed for tabular data on
datasets that did not appear in the original papers.
We want to know whether XGBoost still has the lead on
these medium-sized datasets.

{\bf Methods Compared.}
We compare Hopular, NPTs, XGBoost, CatBoost, and LightGBM.
NPTs are used in a non-transductive setting for a fair comparison.

{\bf Hyperparameter Selection.} 
All hyperparameters are selected on seperate validation sets. For NPTs we perform hyperparameter search as in Table~\ref{tab:hyp_npt}. This includes the hyperparameters that have already been successfully used in \citep{Kossen:21} on small- and medium-sized tabular datasets.
This selection also serves as a constraint on the computational resources invested for Hopular.
For XGBoost, CatBoost, and LightGBM, we apply the same Bayesian hyperparameter optimization 
procedure as described in~\citep{ShwartzZiv:21}. For LightGBM we use the default hyperparameter ranges as specified by \texttt{hyperopt-sklearn}~\citep{Komer:14}.
Section~\ref{sec:hyperparameter_selection} of the Appendix describes the hyperparameter selection in more detail.

{\bf Datasets.}
We select the datasets and dataset splits of \citep{ShwartzZiv:21}, 
where XGBoost performs better than Deep Learning methods that have been
designed for tabular data.
We extend this selection by two datasets for regression: (a) \textit{colleges} was already
used for other Deep Learning methods for tabular data~\citep{Somepalli:21}, and
(b) \textit{sulfur} is publicly available and fits with its 10,082 instances well into
the existing collection of medium-sized datasets.
Table~\ref{tab:medium-sized datasets} gives an overview of the medium-sized datasets.
A detailed description of the datasets can be found in Section~\ref{sec:dataset_explanation} of the Appendix.

\begin{table}[h]
\centering
    \caption{Results of all compared methods on the subset of medium-sized tabular datasets~\citep{ShwartzZiv:21}. For classification tasks (\texttt{C}), the {\em accuracy} is reported. For regression tasks (\texttt{R}), the {\em mean squared error} multiplied by a factor of $1000$ is reported. The reported deviations are the corresponding {\em standard error of the mean}. All values are computed on the respective test sets, averaged over {\em three}~replicates.\label{tab:intel_experiments_results}}
    \begin{center}
        \begin{tabular}{
        l
        S[table-format=2.2(1),separate-uncertainty]
        S[table-format=2.2(1),separate-uncertainty]
        S[table-format=2.2(1),separate-uncertainty]
        S[table-format=2.2(1),separate-uncertainty]
        S[table-format=2.2(1),separate-uncertainty]}
            {\bf Dataset} & {\bf Hopular} & {\bf NPTs} & {\bf XGBoost} & {\bf CatBoost} & {\bf LightGBM} \\
            \toprule
            sulfur (\texttt{R})    &  1.04(02) &  1.24(02) &  1.23(00) &  1.06(01) &  1.16(01) \\
            colleges (\texttt{R})  & 21.18(09) & 25.67(23) & 30.47(00) & 26.40(09) & 25.64(09) \\
            eye (\texttt{C})       & 53.56(48) & 53.21(12) & 57.43(00) & 56.35(05) & 57.34(28) \\
            gesture (\texttt{C})   & 71.20(19) & 67.83(06) & 68.05(00) & 68.86(21) & 69.01(09) \\
            blastchar (\texttt{C}) & 80.05(11) & 79.98(11) & 76.78(00) & 80.13(12) & 79.92(21) \\
            shrutime (\texttt{C})  & 86.12(09) & 85.62(07) & 84.58(00) & 86.39(04) & 86.18(02)
        \end{tabular}
    \end{center}
\end{table}

\textbf{Results.} Table~\ref{tab:intel_experiments_results} reports the results
of Hopular, NPTs, XGBoost, CatBoost, and LightGBM on the medium-sized datasets.
The evaluation procedure is from \citep{ShwartzZiv:21}.
Hopular is the best performing method on 3 out of the 6 datasets.
The runner-up method, CatBoost, is twice the best method, whereas XGBoost once.
The biggest performance difference is achieved by Hopular on the two regression datasets,
where the capabilities of an external memory really shine.
Directly deriving the underlying function for regression datasets may be a difficult task,
especially in absence of abundant data.
Hopular is able to mitigate this shortcoming
by incorporating local neighbourhood information and
iteratively refining its current prediction by memory lookups.
Over the 6 datasets, NPTs and XGBoost have a median rank of 4.5,
CatBoost and LightGBM of 2.5 and 2, respectively,
and Hopular has a median rank of 1.5.
{\bf On average over all 6 datasets, Hopular performs better than 
NPTs, XGBoost, CatBoost, and LightGBM.}
We also found that our method needs only a fraction of the memory compared to NPTs which can be seen in Table~\ref{tab:memory_footprint}. We also added runtime estimates in Table~\ref{tab:training_time}.

\section{Conclusion}
Hopular is a novel Deep Learning architecture where every layer is equipped
with an external memory. This enables Hopular to mimic standard iterative learning
algorithms that refine the current prediction by re-accessing the training set.
We validated the usefulness of this property both on small- and
medium-sized tabular datasets. Hopular is the best performing method
across a broad selection of specifically challenging small-sized UCI
datasets. Additionally, Hopular is the best-performing method on
medium-sized tabular datasets among which CatBoost and LightGBM achieved very competitive
results.
This makes Hopular a strong contender to current state-of-the-art methods like Gradient Boosting and other Deep Learning methods specialized in small- and medium-sized datasets.

\begin{ack}
The ELLIS Unit Linz, the LIT AI Lab, the Institute for Machine Learning, are supported by the Federal State Upper Austria. IARAI is supported by Here Technologies. We thank the projects AI-MOTION (LIT-2018-6-YOU-212), AI-SNN (LIT-2018-6-YOU-214), DeepFlood (LIT-2019-8-YOU-213), Medical Cognitive Computing Center (MC3), INCONTROL-RL (FFG-881064), PRIMAL (FFG-873979), S3AI (FFG-872172), DL for GranularFlow (FFG-871302), AIRI FG 9-N (FWF-36284, FWF-36235), ELISE (H2020-ICT-2019-3 ID: 951847). We thank Audi.JKU Deep Learning Center, TGW LOGISTICS GROUP GMBH, Silicon Austria Labs (SAL), FILL Gesellschaft mbH, Anyline GmbH, Google, ZF Friedrichshafen AG, Robert Bosch GmbH, UCB Biopharma SRL, Merck Healthcare KGaA, Verbund AG, Software Competence Center Hagenberg GmbH, T\"{U}V Austria, Frauscher Sensonic and the NVIDIA Corporation.
\end{ack}

\bibliography{memory}
\bibliographystyle{ml_institute}

\newpage{}
\appendix

\section{Appendix}
\renewcommand\thefigure{\thesection.\arabic{figure}}
\renewcommand\thetable{\thesection.\arabic{table}}

\subsection{Architecture} \label{sec:architecture_appendix}

\begin{figure}[h]
    \centering
    \includegraphics[scale=1.12]{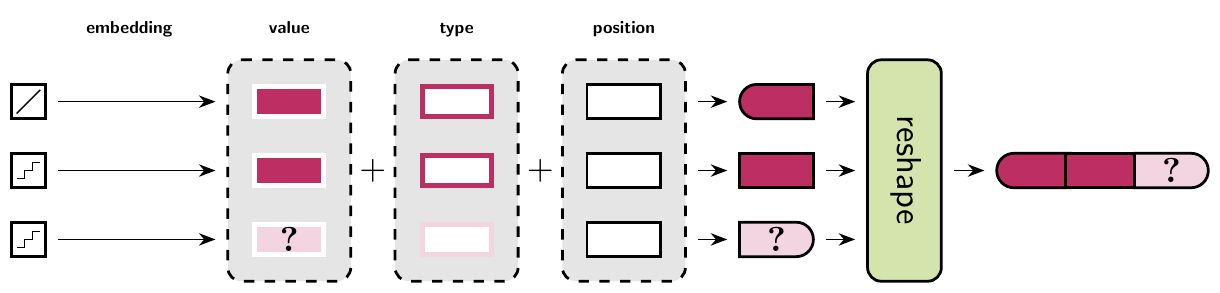}
    \caption{Embedding Layer. All attributes of an original input sample
    are mapped to an $e$-dimensional embedding space. The position of an
    attribute within a sample and the attribute type are conserved by separate $e$-dimensional embeddings. All
    three embedding vectors are summed and serve as the final
    representation of an input attribute. The input sample is represented
    by the concatenation of all its attribute representations.}
    \label{fig:hopular_embedding}
\end{figure}

\begin{figure}[h]
    \centering
    \includegraphics[scale=1.12]{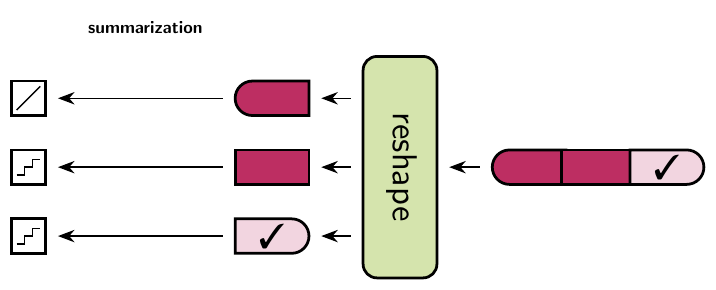}
    \caption{Summarization Layer. The current prediction vector on the right is mapped to the final prediction vector on the left by separately mapping each current attribute
    prediction to its respective final prediction. This final prediction vector lives
    in the same space as the original input sample and is used for the 
    computation of the respective losses.}
    \label{fig:hopular_summarization}
\end{figure}

\subsection{Datasets} \label{sec:dataset_explanation}
\subsubsection{UCI Dataset Selection}
To assess the performance of Hopular and other Deep Learning methods 
on small datasets, 
we select a subset of 21 datasets from \citep{Klambauer:17}. 
The sizes of these datasets range from 200 to 1,000 samples.
We put the focus on smaller sizes, therefore we select
13 datasets with 500 samples or less.
Additionally, we select four datasets with 500 to 750 samples and 
four dataset with 750 to 1,000 samples. 
Small datasets typically have small test sets, 
which introduce a high variance in their evaluations.
This is especially true if they are overly small or unbalanced.
Furthermore, some test sets seem to be not sampled iid from the whole population. 
Thus, the method evaluation may be highly dependent on the chosen train/test split and
performance estimates may be skewed.
Problematic datasets in \citep{Klambauer:17} are characterized
by having a range of accuracy values across well established methods of greater or equal $0.5$
We exclude the problematic datasets 
{\em seeds}, {\em spectf}, {\em libras}, {\em dermatology}, {\em arrythmia}, 
and {\em conn-bench-vowel-deterding}.
The dataset {\em spect} is excluded as its description in \citep{Fernandez:14} 
is in conflict with the available UCI version regarding the number of attributes and samples.
The dataset
{\em heart-hungarian} is excluded as the dataset description 
is insufficient to distinguish between categorical and continuous attributes,
which is required by some methods.
Since {\em breast-cancer-wisc} is practically solved ($0.9859$ accuracy), it is excluded as
it does not allow to distinguish the performances of the compared methods.
We drop {\em heart-va}, 
since the best reported method has only a low accuracy of $0.4$.

\pagebreak{}
\subsubsection{Small-Sized Dataset Description}

\begin{table}[ht]
    \caption{Overview of small-sized datasets with their number of instances, number
    of continuous features, and number of categorical features. All small-sized datasets are classification~tasks.\label{tab:small-sized datasets}}
    \begin{center}
        \begin{tabular}{lS[table-format=4.0]S[table-format=2.0]S[table-format=2.0]}
               {\bf Dataset} & {\makecell{\bf Size\\($N$)}} & {\makecell{\bf \# cont.\\\bf features}} & {\makecell{\bf \# cat.\\\bf features}}\\
               \toprule
             conn-bench & 208 & 60 & 0 \\
             glass & 214 & 9 & 0 \\
             statlog-heart & 270 & 6 & 7 \\
             breast-cancer & 286 & 0 & 9 \\
             heart-cleveland & 303 & 6 & 9 \\
             haberman-survival & 306 & 3 & 0 \\
             vertebral-column2 & 310 & 6 & 0 \\
             vertebral-column3 & 310 & 6 & 0 \\
             primary-tumor & 330 & 0 & 17 \\
             ecoli & 336 & 5 & 0 \\
             horse-colic & 368 & 8 & 19 \\
             congressional-voting & 435 & 0 & 16 \\
             cylinder-bands & 512 & 20 & 19 \\
             monks-2 & 601 & 6 & 0 \\
             statlog-australian-credit & 690 & 5 & 9 \\
             credit-approval & 690 & 6 & 9 \\
             blood-transfusion & 748 & 4 & 1 \\
             energy-y2 & 768 & 7 & 0 \\
             mammographic & 961 & 1 & 5 \\
             led-display & 1000 & 0 & 6 \\
             statlog-german-credit & 1000 & 23 & 0
        \end{tabular}
    \end{center}
\end{table}

Below we give more precise descriptions of the datasets used in our small-sized experiments:
\begin{description}[topsep=-1.5pt,itemsep=1.5pt]
    \item {\em conn-bench-sonar-mines-rocks} or  {\em Connectionist Bench (Sonar, Mines vs. Rocks)}: A classification setting of 208 instances with 60 continuous features per instance. The task is to discriminate between sonar sounds from metal vs. rocks.
    
    \item {\em glass} or {\em Glass Identification}: A classification setting of 214 instances with 9 continuous features per instance. The task is to discriminate between 6 types of glass.
    
    \item {\em statlog-heart}: A classification setting of 270 instances with 6 continuous and 7 categorical features per instance. The task is to predict the presence or absence of a heart disease.
    
    \item {\em breast-cancer}: A classification setting of 286 instances with 9 categorical features per instance. The task is to predict the presence or absence of breast cancer.
    
    \item {\em heart-cleveland} or {\em Heart Disease}: A classification setting of 303 instances with 6 continuous and 7 categorical features per instance. The task is to predict the presence or absence of a heart disease.
    
    \item {\em haberman-survival}: A classification setting of 306 instances with 3 continuous features per instance. The task is to predict whether patients survived longer than 5 years or not.
    
    \item {\em vertebral-column2, vertebral-column3} or {\em Vertebral Column Dataset}: Two classification settings of 310 instances each with 6 continuous features per instance. The task is to classify patients into either 2 or 3 classes.
    
    \item {\em primary-tumor}: A classification setting of 330 instances with 17 categorical features per instance. The task is to predict the class of primary tumors.
    
    \item {\em ecoli}: A classification setting of 336 instances with 5 continuous and 2 categorical features per instance. The tasks is to classify proteins into 8 classes.
    
    \item {\em horse-colic}: A classification setting of 368 instances with 8 continuous and 19 categorical features per instance. The task is to predict the survival or death of a horse.
    
    \item {\em congressional-voting}: A classification setting of 435 instances with 16 categorical features per instance. The task is to predict political affiliation.
    
    \item {\em cylinder-bands}: A classification setting of 512 instances with 20 continuous and 19 categorical features per instance. The task is to classify the band type.
    
    \item {\em credit-approval}: A classification setting of 690 instances with 6 continuous and 9 categorical features per instance. The task is to determine positive or negative feedback for credit card applications.
    
    \item {\em blood-transfusion} or {\em Blood Transfusion Service Center}: A classification setting of 748 instances with 4 continuous and 1 categorical feature per instance. The task is to predict whether a person donated blood or not.
    
    \item{\em statlog-german-credit}: A classification setting of 1,000 instances with 23
    continuous features per instance. The goal is to determine credit-worthiness of customers.
    
    \item {\em mammographic} or {\em Mammographic Mass}: A classification setting of 961 instances with 1 continuous and 5 categorical features per instance. The task is to discriminate between benign and malignant mammographic masses.
    
    \item{\em led-display}: A classification setting of 1,000 instances with 6 categorical features
    per instance. The task is to classify decimal digits from light-emiting diodes with noise.
    
    \item {\em statlog-australian-credit}: A classification setting of 690 instances
    with 5 continuous and 9 categorical features. The task to grant customers
    credit-approval or not.
    
    \item {\em energy-y2} or {\em Energy efficiency Data Set}:
    A classification setting of 768 instances with 7 continuous
    features per instance. The task is to predict the cooling load for a given building.
    
    \item {\em monks-2} It is part of the {\em Monk's Problems Data Set}. A classification
    task for 601 instances with 6 categorical features. The task is to discriminate
    between two classes.
\end{description}

\subsubsection{Medium-Sized Dataset Description}

\begin{table}[ht]
    \caption{Medium-sized datasets with their number of instances, number
    of continuous features, and number of categorical features. Classification tasks are marked with (\texttt{C}), whereas regression tasks are marked with (\texttt{R}).\label{tab:medium-sized datasets}}
    \begin{center}
        \begin{tabular}{lS[table-format=5.0]S[table-format=2.0]S[table-format=2.0]}
               {\bf Dataset} & {\makecell{\bf Size\\($N$)}} & {\makecell{\bf \# cont.\\\bf features}} & {\makecell{\bf \# cat.\\\bf features}}\\
               \toprule
             blastchar (\texttt{C}) & 7048 & 3 & 17 \\
             colleges (\texttt{R}) & 7064 & 33 & 12 \\
             gesture-phase (\texttt{C}) & 9873 & 31 & 0 \\
             shrutime (\texttt{C}) & 10000 & 2 & 9 \\
             sulfur (\texttt{R}) & 10082 & 5 & 0 \\
             eye-movements (\texttt{C}) & 10936 & 19 & 3
        \end{tabular}
    \end{center}
\end{table}

Below we give more precise descriptions of the datasets used in our medium-sized experiments:
\begin{description}[topsep=-1.5pt,itemsep=1.5pt]
     \item {\em shrutime}: A classification setting of 10,000 instances with 2 continuous and 9 categorical features per instance. The task is to predict whether a bank account is closed or not.
    
    \item {\em blastchar}: A classification setting of 7,048 instances with 3 continuous and 17 categorical features per instance. The task is to predict customer behavior.
    
    \item {\em gesture} or {\em gesture-phase} or {\em Gesture Phase Segmentation}: A classification setting of 9,873 instances with 31 continuous features per instance. The task is to classify gesture phases.
    
    \item {\em eye} or {\em eye-movements}: A classification setting of 10,936 instances with 19 continuous and 3 categorical features per instance. The task is to discriminate between correct, irrelevant or relevant answers.
    
    \item {\em colleges}: A regression setting of 7,064 instances with 33 continuous and
    12 categorical features per instance. The task is to predict pell grant percentages
    for colleges in the USA.
     
    \item {\em sulfur}: A regression setting of 10,082 instances with 5 continuous features
    per instance. The task is to predict H2S concentration in a factory module.
\end{description}

\subsection{Hyperparameter selection process}\label{sec:hyperparameter_selection}
\begin{table}[ht]
    \caption{Complete listing of all evaluated hyperparameter settings for NPTs. For all experiments a learning rate of $0.001$ as well as a dropout probability of $0.1$ is used. Settings marked with an asterisk (*) are not performed on \textit{conn-bench-sonar-mines-rocks} due to out-of-memory issues.\label{tab:hyp_npt}}
    \begin{center}
        \begin{tabular}{cS[table-format=2.0]S[table-format=2.0]S[table-format=1.1]S[table-format=1.2]cS[table-format=4.0]c}
            {\bf \makecell{dataset\\group}} & {\bf \makecell{\# netw.\\layers}} 
            & {\bf \makecell{\# att.\\heads}} & {\bf \makecell{label mask.\\prob.}}
            & {\bf \makecell{feature mask.\\prob.}} & {\bf \makecell{learn. rate\\scheduler}}
            & {\bf \makecell{emb.\\dim.}} & {}
            \\
            \toprule \\
            \multirow{8}{*}{\makecell{\it small and\\\it medium}} & 8 & 8 & 1.0 & 0.15 & cosine & 32 &\\
            &16 & 8 & 1.0 & 0.15 & cosine & 32 &\\
            &8 & 16 & 1.0 & 0.15 & cosine & 32 &\\
            &16 & 16 & 1.0 & 0.15 & cosine & 32 &\\
            &8 & 8 & 0.1 & 0.15 & cosine & 32 &\\
            &8 & 8 & 0.5 & 0.15 & cosine & 32 &\\
            &8 & 8 & 1.0 & 0.20 & cosine & 32 &\\
            &8 & 8 & 1.0 & 0.15 & cosine cyclic & 32 &\\ \midrule
            \multirow{8}{*}{\makecell{\it small}} & 8 & 8 & 1.0 & 0.15 & cosine & 128 &\\
            &16 & 8 & 1.0 & 0.15 & cosine & 128 & *\\
            &8 & 16 & 1.0 & 0.15 & cosine & 128 &\\
            &16 & 16 & 1.0 & 0.15 & cosine & 128 & *\\
            &8 & 8 & 0.1 & 0.15 & cosine & 128 &\\
            &8 & 8 & 0.5 & 0.15 & cosine & 128 &\\
            &8 & 8 & 1.0 & 0.20 & cosine & 128 &\\
            &8 & 8 & 1.0 & 0.15 & cosine cyclic & 128 &
        \end{tabular}
    \end{center}
\end{table}

For the hyperparameter selection process for NPTs we follow \citep{Kossen:21} and take exactly
the same hyperparameter settings that were successfully used among several
datasets. We use these hyperparameter settings for experiments on small- and medium-sized
datasets. For small-sized datasets we additionally use these settings with an increased
embedding dimension of $128$. Especially for such datasets the discrimination among similar samples can be a challenging task. This problem can be mitigated 
by mapping to a higher-dimensional embedding space where the samples have greater
distances between each other.
NPTs follow a masking procedure similar to \citep{Devlin:19}
which is realized by feature and label masking probabilities.
Following the strategy in \citep{Kossen:21} we use the LAMB~\citep{You:20} optimizer for all NPT experiments, extended by a Lookahead~\citep{Zhang1:19} wrapper with fixed values.
For LAMB we use $\beta_L = (0.9, 0.999)$, $\epsilon = 1e{-6}$ and for Lookahead $\alpha = 0.5$, $k = 6$.
The hyperparameter settings for NPTs are shown in Table~\ref{tab:hyp_npt}.

\begin{table}[ht]
    \caption{Complete listing of all evaluated hyperparameter settings for Hopular. For all experiments a learning rate of $0.001$ was used. The dropout probabilities $p_{i}$, $p_{h}$ and $p_{o}$ refer to the embedding layer, Hopular Block and summarization layer, respectively. The three settings of the second group (\textit{medium-sized}) were performed in a non-exhaustive way w.r.t. to all medium-sized datasets.\label{tab:hyp_hop}}
    \begin{center}
        \begin{tabular}{
        c
        S[table-format=1.0]
        S[table-format=2.0]
        c
        S[table-format=1.3]
        S[table-format=1.3]
        S[table-format=1.2]
        S[table-format=1.2]
        S[table-format=1.2]
        S[table-format=1.1]}
            {\multirow{2}{*}{\bf \makecell{dataset\\group}}} &
            {\multirow{2}{*}{\bf \makecell{\# Hop.\\blocks}}} &
            {\multirow{2}{*}{\bf \makecell{\# Hop.\\nets}}} &
            {\multirow{2}{*}{\bf \makecell{$\beta$-scaling\\factor}}} &
            {\multirow{2}{*}{\bf \makecell{mask\\prob.}}} &
            {\multirow{2}{*}{\bf \makecell{replace\\prob.}}} &
            {\multirow{2}{*}{\bf \makecell{weight\\decay}}} &
            \multicolumn{3}{c}{\bf dropout} \\ &&&&&&&
            {\makecell{$p_{i}$}} &
            {\makecell{$p_{h}$}} &
            {\makecell{$p_{o}$}}
            \\
            \toprule \\
            \multirow{4}{*}{\makecell{\it small and\\\it medium}} & 4 & 8 & $10^{\left\{0,2,3\right\}}$ & 0.025 & 0.175 & 0.1 & 0.1 & 0.1 & 0.01\\
            &8 & 8 & $10^{\left\{0,2,3\right\}}$ & 0.025 & 0.175 & 0.1 & 0.1 & 0.1 & 0.01\\
            &4 & 16 & $10^{\left\{0,2,3\right\}}$ & 0.025 & 0.175 & 0.1 & 0.1 & 0.1 & 0.01\\
            &8 & 16 & $10^{\left\{0,2,3\right\}}$ & 0.025 & 0.175 & 0.1 & 0.1 & 0.1 & 0.01\\ \midrule
            \multirow{1}{*}{\makecell{\it medium}} & 8 & 16 & $10^{\left\{0\right\}}$ & 0.000 & 0.000 & 0.0 & 0.0 & 0.0 & 0.00
        \end{tabular}
    \end{center}
\end{table}

For a fair comparison we upper bound Hopular's capacity by the capacity of NPTs which results
in the settings shown in Table \ref{tab:hyp_hop}.
As Hopular provides an additional adjustable scaling factor for $\beta$, we also test scaling factors of $100$ and $1000$ to further emphasize nearest-neighbor search.
In our default setting the weighting term $\gamma$ for our objective in Eq.~\eqref{eq:total_loss}
is annealed using a cosine scheduler starting at $1$ with a final value of $0$.
For medium-sized datasets we also perform experiments with an initial $\gamma$ value of $0.5$. 
We use the original BERT masking as in \citep{Devlin:19}.
Since we store the training data in $H_s$ we have to make sure that
the model does not just learn to retrieve the original input sample
from the training set (like a database query). This is why we 
independently of BERT masking always
mask the corresponding sample in the training set.
We use default values for masking and dropout.
For the medium-sized datasets we also test two different settings of weight decay, and of dropout probabilities in the
Embedding layer, Hopular block and Summarization layer.
In contrast to NPTs, we always mask all labels.
In our experiments the Hopfield dimension $h$ (as described in Section \ref{sec:mhn_tabular})
is fixed by the embedding size $e$, the number of features $d$ and
the number of Hopfield networks $M$ such that $h = d \cdot e / M$.
The LAMB~\citep{You:20} optimizer is used for all Hopular experiments, extended by a method similar to Lookahead~\citep{Zhang1:19} but without synchronization of
fast and slow weights. This is analogous to the exponential moving average used in
\citep{Grill:20}.
For LAMB we use $\beta_L = (0.9, 0.999)$, $\epsilon = 1e{-6}$ and for Lookahead $\alpha = 0.005$, $k = 1$.
NPTs and Hopular are both trained for 10,000 epochs with
early stopping.

For XGBoost and CatBoost we use the package  \texttt{hyperopt}
and apply the same Bayesian hyperparameter optimization procedure
as described in \cite{ShwartzZiv:21}. For all Boosting methods we 
thereby evaluate 1,000 different hpyerparameter settings. More precisely,
the hyperparameters and their search spaces for XGBoost are defined in the following.
\begin{itemize}
    \item \textit{Learning rate:} Log-Uniform distribution $[-7, 0]$
    \item \textit{Max depth:} Discrete uniform distribution $[1, 10]$
    \item \textit{Subsample:} Uniform distribution $[0.2, 1]$
    \item \textit{Colsample bytree:} Uniform distribution $[0.2, 1]$
    \item \textit{Colsample bylevel:} Uniform distribution $[0.2, 1]$
    \item \textit{Min child weight:} Log-Uniform distribution $[-16, 2]$
    \item \textit{Alpha:} Uniform choice $\{0, \text{Log-Uniform }[-16, 2] \}$
    \item \textit{Lambda:} Uniform choice $\{0, \text{Log-Uniform }[-16, 2] \}$
    \item \textit{Gamma:} Uniform choice $\{0, \text{Log-Uniform }[-16, 2] \}$
    \item \textit{Number of estimators:} $1000$
\end{itemize}

It is important to mention that the package \texttt{hyperopt}
defines the Log-Uniform distribution by the
exponents of the respective interval boundaries~\---{}~e.g. 
$\text{Log-Uniform}[-7, 0]$ is defined on $[e^{-7}, e^0]$.
The hyperparameters and their search spaces for CatBoost are defined in the following.
\begin{itemize}
    \item \textit{Learning rate:} Log-Uniform distribution $[-5, 0]$
    \item \textit{Random strength:} Discrete uniform distribution $[1, 20]$
    \item \textit{Max size:} Discrete uniform distribution $[0, 25]$
    \item \textit{L2 leaf regularization:} Log-Uniform distribution $[ \log1, \log10 ]$
    \item \textit{Bagging temperature:} Uniform distribution $[0, 1]$
    \item \textit{Leaf estimation iterations:} Discrete uniform distribution $[1, 20]$
    \item \textit{Number of estimators:} $1000$
\end{itemize}
For LightGBM we use the default hyperparameter ranges as specified by \texttt{hyperopt-sklearn}~\citep{Komer:14}.
\begin{itemize}
    \item \textit{Learning rate:} Log-Uniform distribution $[\log 0.0001, \log 0.5] - 0.0001$
    \item \textit{Max depth:} Discrete uniform distribution $[1, 11]$
    \item \textit{Number of leaves:} Discrete uniform distribution $[2, 121]$
    \item \textit{Gamma:} Log-Uniform distribution $[\log 0.001, \log 5] - 0.0001$
    \item \textit{Min child weight:} Log-Uniform distribution $[\log 1, \log 100]$
    \item \textit{Subsample:} Uniform distribution $[0.5, 1]$
    \item \textit{Colsample bytree:} Uniform distribution $[0.5, 1]$
    \item \textit{Colsample bylevel:} Uniform distribution $[0.5, 1]$
    \item \textit{Alpha:} Log-Uniform distribution $[\log 0.0001, \log 1]$
    \item \textit{Lambda:} Log-Uniform distribution $[\log 1, \log 4]$
    \item \textit{Boosting type:} Uniform choice $\{ \text{gbdt, dart, goss}\}$
    \item \textit{Number of estimators:} $1000$
\end{itemize}

\subsection{Results}
In Table \ref{tab:uci_experiments_complete} we show the median rank across
all 21 selected UCI datasets. Methods are ranked for each dataset
according to their accuracy on the respective test set.
\begin{table}[ht]
    \caption{Median rank of compared methods across 
    the datasets of the UCI machine learning repository. 
    Methods are ranked for each dataset according to the accuracy on the respective test set.
    Hopular achieves the lowest median rank of $7.5$, therefore is the best
    performing method across the considered UCI datasets. \label{tab:uci_experiments_complete}}
    \begin{center}
        \begin{tabular}{lS[table-format=2.1]lS[table-format=2.1]}
            {\bf Method} & {\bf Rank} & {\bf Method} & {\bf Rank} \\
            \toprule
            Hopular (DL)                        &  7.5 & Rule-Based Methods        & 15.0 \\
            Support Vector Machines             &  9.5 & Other Ensembles           & 15.0 \\
            Logistic and Multinomial Regression & 10.0 & BatchNorm (DL)            & 15.0 \\
            Random Forest                       & 11.0 & Boosting Methods          & 15.0 \\
            Self-Normalizing Networks (DL)      & 11.0 & Generalized Linear Models & 15.5 \\
            Non-Parametric Transformers (DL)    & 11.0 & WeightNorm (DL)           & 15.5 \\
            Neural Networks (DL)                & 11.5 & Discriminant Analysis     & 16.0 \\
            XGBoost                             & 12.0 & Other Methods             & 17.5 \\
            Multivariate Adaptive Reg. Splines  & 12.0 & ResNet (DL)               & 19.0 \\
            Decision Trees                      & 13.5 & LayerNorm (DL)            & 19.0 \\
            MSRAinit (DL)                       & 14.0 & Partial Least Squares     & 19.5 \\
            Bagging Methods                     & 14.0 & Bayesian Methods          & 20.0 \\
            CatBoost                            & 14.0 & Nearest Neighbour         & 24.0 \\
            LightGBM                            & 14.5 & Stacking (Wolpert)        & 28.0 \\
            Highway Networks (DL)               & 14.5 &                           &
        \end{tabular}
    \end{center}
\end{table}

\subsection{Memory footprint and runtime estimates}
In table~\ref{tab:memory_footprint} we show the memory footprint of Hopular and NPTs for all medium-sized datasets ranging from the smallest to the largest model. In all cases the whole training set is stored in the memory of module $H_s$. Even in the full batch setting where all the data is used as model input there is no prohibitive memory increase. In contrast, NPTs have a much higher memory memory consumption in the full batch setting. There, for 3 datasets the larger models even run out of memory on an Nvidia A100 GPU.
\begin{table}[h]
    \caption{Memory footprint of Hopular and NPTs in \emph{gibibytes (GiB)} for medium-sized datasets ranging from our smallest to largest model. Settings with a memory footprint of $80.00\text{{\raisebox{.2ex}+}}$ are not performed due to out-of-memory issues. \label{tab:memory_footprint}\vspace{0.5\baselineskip}}
    \begin{center}
        \begin{tabular}{l
        S[table-format=2.2]@{\hskip 1mm}
        c@{\hskip -2mm}
        S[table-format=2.2]
        S[table-format=2.2]@{\hskip 1mm}
        c@{\hskip 0.5mm}
        S[table-format=2.2]
        S[table-format=2.2]@{\hskip 1mm}
        c@{\hskip -2mm}
        S[table-format=2.2]
        S[table-format=2.2]@{\hskip 1mm}
        c@{\hskip 1.0mm}
        S[table-format=2.2]}
               {\multirow{2}{*}{\bf Dataset}} & \multicolumn{6}{c}{\bf Hopular} & \multicolumn{6}{c}{\bf NPTs} \\
               & \multicolumn{3}{c}{\bf single sample} & \multicolumn{3}{c}{\bf full batch} & \multicolumn{3}{c}{\bf single sample} & \multicolumn{3}{c}{\bf full batch} \\
               \toprule
             blastchar (\texttt{C})     & 2.38 & to & 2.75 &  4.83 & to &  7.61
                                        & 1.97 & to & 2.38 & 20.49 & to & 56.17 \\
             colleges (\texttt{R})      & 3.13 & to & 3.90 &  6.58 & to & 11.62
                                        & 3.98 & to & 6.09 & 27.13 & to & 74.56 \\
             gesture-phase (\texttt{C}) & 2.77 & to & 3.41 &  8.92 & to & 15.61
                                        & 2.73 & to & 3.90 & 40.95 & to & 80.00{\raisebox{.2ex}+} \\
             shrutime (\texttt{C})      & 2.60 & to & 3.23 &  7.53 & to & 13.05
                                        & 1.66 & to & 1.79 & 36.30 & to & 78.75 \\
             sulfur (\texttt{R})        & 2.55 & to & 3.18 &  7.54 & to & 13.14
                                        & 1.55 & to & 1.59 & 35.95 & to & 80.00{\raisebox{.2ex}+} \\
             eye-movements (\texttt{C}) & 2.68 & to & 3.28 & 10.19 & to & 18.21
                                        & 2.11 & to & 2.67 & 45.92 & to & 80.00{\raisebox{.2ex}+}
        \end{tabular}
    \end{center}
\end{table}

In table~\ref{tab:training_time} we perform measurements on training and inference times.
We show the step time for medium-sized datasets during training. Inference times are assumed to be much lower, as no gradient computation and parameter updates need to be performed.

\begin{table}[h]
    \caption{Step time of Hopular and NPTs in \emph{milliseconds (ms)} during training.\label{tab:training_time}\vspace{0.5\baselineskip}}
    \begin{center}
        \begin{tabular}{l
        S[table-format=3.2(2),separate-uncertainty]
        S[table-format=4.2(2),separate-uncertainty]
        S[table-format=3.2(2),separate-uncertainty]
        S[table-format=4.2(2),separate-uncertainty]}
               {\multirow{2}{*}{\bf Dataset}} & \multicolumn{2}{c}{\bf Hopular} & \multicolumn{2}{c}{\bf NPTs}\\
               & {\bf single sample} & {\bf full batch} & {\bf single sample} & {\bf full batch}\\
               \toprule
             blastchar (\texttt{C})     &  73.69(02) &  503.45(08)
                                        &  81.74(11) &  167.26(25) \\
             colleges (\texttt{R})      & 120.15(09) &  824.34(17)
                                        & 118.13(13) &  321.32(25) \\
             gesture-phase (\texttt{C}) &  95.40(03) & 1155.47(06)
                                        &  99.38(08) &  384.58(16) \\
             shrutime (\texttt{C})      &  61.90(02) &  652.81(04)
                                        &  68.18(08) &  182.11(16) \\
             sulfur (\texttt{R})        &  52.71(02) &  629.55(04)
                                        &  59.44(08) & 159.86(28) \\
             eye-movements (\texttt{C}) &  76.94(02) & 1141.37(03)
                                        &  84.21(08) & 338.53(18)
        \end{tabular}
    \end{center}
\end{table}

\subsection{Hopular Intuition: Mimicking Iterative Learning}\label{sec:iter_learn}
In our first example we consider Nadaraya-Watson kernel regression
\citep{Watson:64,Nadaraya:64,Benedetti:77,Weinberger:07}.
The training set is
$\{(\Bz_1,\By_1),\ldots,(\Bz_N,\By_N)\}$ 
with inputs $\Bz_i$ summarized by the input
matrix $\BZ = (\Bz_1,\ldots,\Bz_N)$ and labels $\By_i$ summarized
in the label matrix $\BY=(\By_1,\ldots,\By_N)$. The kernel function 
is $k(\Bz_i,\Bz)$.
The estimator $\Bg$ for $\By$ given $\Bz$ is:
\begin{align}
    \Bg(\Bz) \ &= \  \sum_{i=1}^N \By_i \ \frac{k(\Bz_i,\Bz)}{\sum_{i=1}^N  k(\Bz_i,\Bz)} \ .
\end{align}

By using the RBF kernel we get:
\begin{align}
k(\Bz_i,\Bz_j) \ &= \  \exp(- \ \beta/2  \ \NRM{\Bz_i \ - \ \Bz_j}^2 ) \ =
 \  \exp(- \ \beta/2  \ ( \Bz_i^T \Bz_i \ - \ 2 \ \Bz_i^T \Bz_j \ + \
                   \Bz_j^T \Bz_j ) )\ .
\end{align}
For normalized vector $\Bz_i$ we have $ \Bz_i^T \Bz_i = \NRM{\Bz_i}^2
=1$, therefore
\begin{align}
  k(\Bz_i,\Bz_j) \ &= \   \exp(- \ \beta  \ ( 1 \ - \ \Bz_i^T \Bz_j) )
   \ = \ c \  \exp( \beta  \ \Bz_i^T \Bz_j )\ .
\end{align}
We obtain for Nadaraya–Watson kernel regression  with the RBF kernel and normalized inputs:
\begin{align} \label{eq:nadwat}
\Bg(\Bz) \ &= \   \BY \ \soft(\beta \ \BZ^T \ \Bz)  \ .
\end{align}

Metric learning for kernel regression learns the kernel $k$
which is the distance function \citep{Weinberger:07}. A Hopular Block can
do the same in Eq.~\ref{eq:Hs} via learning the weight matrices $\BW_{\BX}$
and $\BW_{\Bxi}$. If we set in Eq.~\ref{eq:nadwat}:
\begin{align}
    \BZ^T = \BX^T \ \BW_{\BX}^T, \ \ \ \Bz = \BW_{\Bxi} \ \Bxi, \ \ \ \BY = \BW_{\BS} \ \BW_{\BX} \ \BX
\end{align}
then we obtain Eq.~\ref{eq:Hs}, with the fixed label matrix $\BY$. \\

In the second example we show how Hopular can realize a linear model
with the AdaBoost Objective. The AdaBoost objective for classification
with a binary target $y \in \{ -1, +1 \}$ can be written as follows~\---{}~see Eq.~3 and Eq.~4 in \citep{Shen:10}:
\begin{align}
    \rL \ &= \  \ln \sum_{i=1}^{N}  \exp(- \ y_i \ g(\Bz_i) ) \ . 
\end{align}
We use this objective for learning the linear model:
\begin{align}
    g(\Bz_i) \ &= \ \beta \ \Bxi^T \Bz_i \ . 
\end{align}
The objective multiplied by $\beta^{-1}$ with $\BY$ as the diagonal matrix
of the targets $\{ \By_1, \cdots , \By_N \}$ becomes:
\begin{align}
  \rL \ &= \  \beta^{-1} \ \ln \sum_{i=1}^{N}  \exp(- \ \beta \ y_i \ \Bxi^T \Bz_i ) \ 
  = \ \mathrm{lse}(\beta \ , \ - \ \BY \ \BZ^T \  \Bxi)  \ , 
\end{align}
where $\mathrm{lse}$ is the log-sum-exponential function.
The gradient of this objective is:
\begin{align}
  \frac{\partial \rL}{\partial \Bxi} \ &= \  
  - \ \BZ \ \BY \ \soft( - \ \beta \ \BY \ \BZ^T \  \Bxi )  \ . 
\end{align}
This is Eq.~\ref{eq:Hs} with:
\begin{align}
    \BY \ \BZ^T = \BX^T \ \BW_{\BX}^T, \ \ \ \BW_{\Bxi} = \BI, \ \ \ \BW_{\BS} = \BI
\end{align}
Thus, a Hopular Block can implement a gradient descent update rule for a linear
classification model using the AdaBoost objective function. The current
prediction $\Bxi$ comes from the previous layer.

These are two additional examples among the standard iterative learning
algorithms which Hopular can mimic.

\subsection{Source code}
Source code is available at: \url{https://github.com/ml-jku/hopular}

\end{document}